\documentclass{article}

\usepackage[final]{corl_2024} 
\usepackage{graphicx}
\usepackage{amsmath,amssymb}

\usepackage{etoolbox}
\usepackage{longtable}
\usepackage{pdflscape}
\usepackage[svgnames]{xcolor}
\usepackage{colortbl}
\usepackage{booktabs}
\usepackage{caption}

\makeatletter

\newcommand{\reducedplus}{\mathpalette\reduced@plus\relax}
\newcommand{\reduced@plus}[2]{%
  \sbox6{$\m@th#1+$}%
  \sbox8{\scalebox{0.875}{\copy6}}%
  \dimen@=\dimexpr(\wd6-\wd8)/3\relax
  \raisebox{\dimen@}{\box8}%
} 

\def\tabref#1{Tab.~\ref{#1}}
\def\eqref#1{Eq.~(\ref{#1})}

\title{Multi-Modal 3D Scene Graph Updater for Shared and Dynamic Environments}

\author{
  Emilio Olivastri\\
  Department of Information Engineering\\
  University of Padova \\
  Italy\\
  \texttt{olivastrie@dei.unipd.it} \\
  \And
  Jonathan Francis \\
  Bosch Center for AI \\
  Carnegie Mellon University\\
  United States\\
  \texttt{jon.francis@us.bosch.com} \\
  \And
  Alberto Pretto \\
  Department of Information Engineering \\
  University of Padova\\
  Italy\\
  \texttt{alberto.pretto@dei.unipd.it} \\
  \And
  Niko S\"underhauf \\
  Centre for Robotics \\
  Queensland University of Technology \\
  Australia \\
  \texttt{niko.suenderhauf@qut.edu.au}
 \And
  Krishan Rana \\
  Centre for Robotics \\
  Queensland University of Technology \\
  Australia \\
  \texttt{ranak@qut.edu.au} \\
}

\begin{document}
\maketitle


\begin{abstract}

The advent of generalist Large Language Models (LLMs) and Large Vision Models (VLMs) have streamlined the construction of semantically enriched maps that can enable robots to ground high-level reasoning and planning into their representations.   One of the most widely used semantic map formats is the 3D Scene Graph, which captures both metric (low-level) and semantic (high-level) information. However, these maps often assume a static world, while real environments, like homes and offices, are dynamic. Even small changes in these spaces can significantly impact task performance. To integrate robots into dynamic environments, they must detect changes and update the scene graph in real-time. This update process is inherently multimodal, requiring input from various sources, such as human agents, the robot’s own perception system, time, and its actions. This work proposes a framework that leverages these multimodal inputs to maintain the consistency of scene graphs during real-time operation, presenting promising initial results and outlining a roadmap for future research.
\end{abstract}



\section{Introduction}
	
A fundamental goal in robotics research is to enable robots to operate safely, efficiently, and autonomously in environments shared with humans. In such shared spaces—whether homes, offices, or other collaborative settings—robots must perform a wide range of tasks, from simple object manipulation to complex interactions with the environment. To achieve this, robots rely heavily on internal representations of their surroundings, typically in the form of maps. These maps allow robots to understand spatial layouts and reason about tasks at both low-level (e.g., navigation) and high-level (e.g., task planning). However, one key assumption often made when using these maps is that the environment remains static after the map is created.

In practice, this assumption breaks down in dynamic, shared environments where human and robotic activities constantly alter the state of the world. Furniture is moved, objects are relocated, and items are used or discarded. These changes are non-trivial, and any discrepancies between the robot’s internal map and the real-world environment can significantly degrade task performance. For instance, a robot instructed to retrieve a banana from the kitchen may fail if it relies on outdated map information—such as the banana being present a week ago—without accounting for the possibility that it has since been eaten or thrown away. Such inconsistencies can lead to inefficient behavior, incorrect task execution, and ultimately limit the robot’s usefulness in dynamic environments.

To prevent this degradation in performance, robots must possess the ability to detect changes in their environment and update their internal maps in real time. This capability is essential for ensuring that downstream reasoning and task planning are based on an accurate, up-to-date representation of the world. A critical insight from this work is that maintaining the consistency of a robot’s map in shared, living environments is inherently a multimodal challenge. It requires the robot to integrate information from multiple sources—such as visual perception, interactions with humans and other agents, temporal data, and its own actions—to inform map updates. The ability to leverage this diverse set of information streams is key to building robust, real-time scene representations that evolve as the environment changes.

One of the most prominent representations used in robotics for modeling complex environments is the 3D Scene Graph (3DSG)\citep{armeni20193d}. 3DSGs encode both metric and semantic information, allowing robots to reason about not only the spatial configuration of objects but also their relationships and roles in the environment. The flexibility of 3DSGs makes them a powerful tool for representing dynamic environments, as they can be extended to include different types of information. For example, hierarchical relationships between objects and their surrounding spaces (e.g., an object belonging to a room) can be incorporated\citep{werby2024hierarchical}, as well as semantic relationships between objects (e.g., objects that are typically used together)~\citep{gu2024conceptgraphs}. Despite the versatility of 3DSGs, existing research has primarily focused on their construction and optimization for specific tasks, with limited attention to how these representations can be dynamically updated to reflect ongoing changes in the real world.

To address this critical gap, we propose MM-3DSGU, a general framework designed to dynamically update 3D Scene Graphs in shared, dynamic environments. MM-3DSGU leverages heterogeneous sources of information, including time, human interactions, the robot’s perception system, and its own actions, to perform robust change detection and maintain an accurate representation of the environment. By integrating these multimodal streams, our framework ensures that the robot’s scene graph is consistently aligned with the real world, enabling better task execution and reasoning in dynamic settings.

In this work, we present the design and initial implementation of MM-3DSGU, along with promising results from preliminary experiments. We also outline a detailed roadmap for future research, aimed at further refining our approach and addressing challenges related to scalability, efficiency, and real-time performance in more complex environments.

\begin{figure}[t!]
   \centering
      \centering
   \begin{minipage}[b]{\linewidth}
      \includegraphics[width=\linewidth]{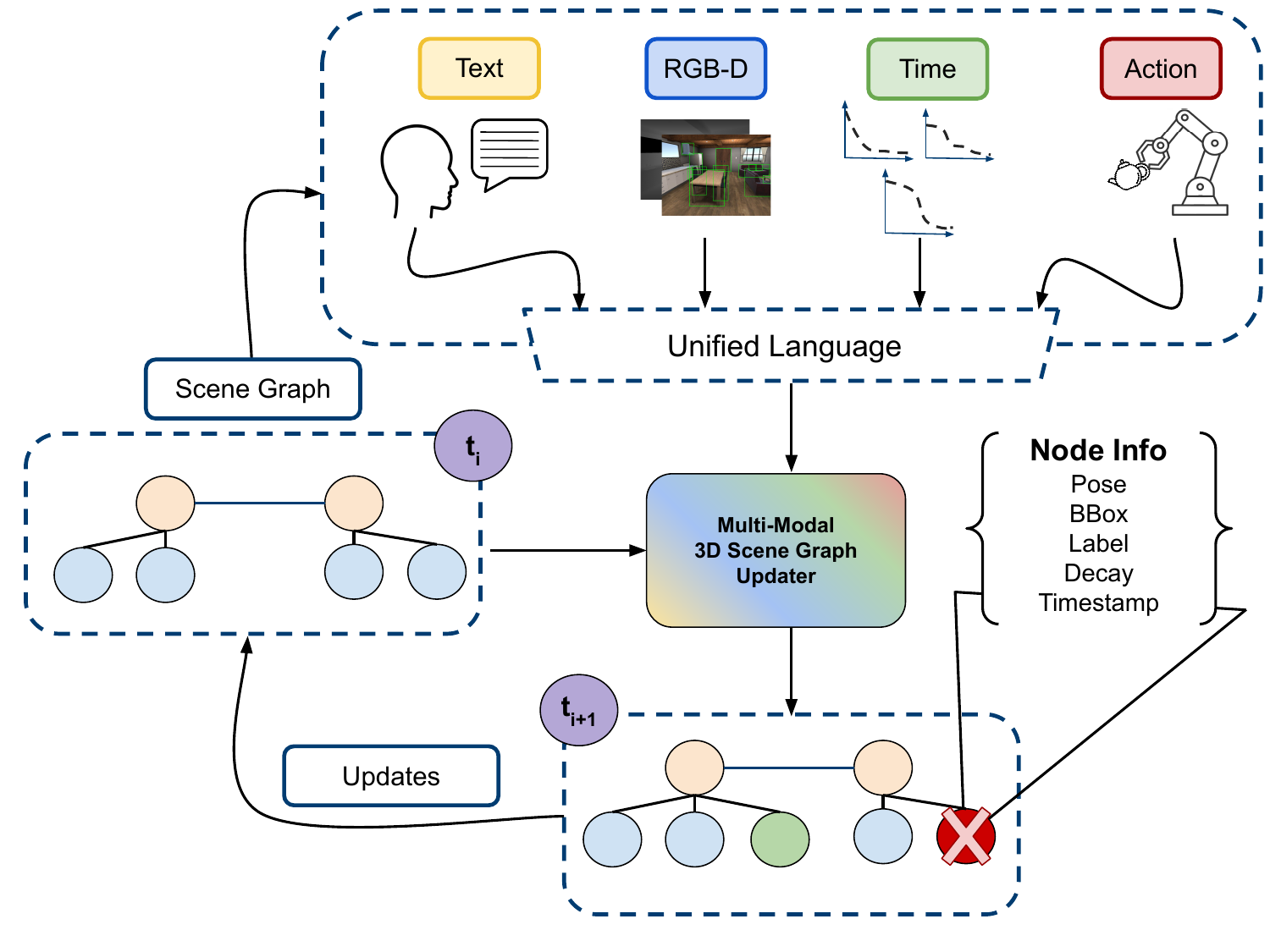}
   \end{minipage}\hfill
   \caption{\textbf{High-level overview of the proposed framework.} The various modalities proposed for change detection are displayed (top). These modules leverage the latest version of the 3D scene graph to assess whether change has occurred. Update operations on the 3D scene graph (bottom) are applied agnostically, due to the use of a unified language.}
   \label{fig:teaser}
\end{figure}

\section{Related Work}
\label{sec:rel_works}

\textbf{3D Scene Graphs}: 3D Scene Graphs (3DSGs)~\citep{armeni20193d} are a powerful representation for modeling large-scale environments by abstracting them into graphs, where nodes represent objects and spatial concepts, and edges capture relationships between them. This structure enables more efficient high-level reasoning for robotic tasks such as navigation and manipulation~\citep{hughes2022hydra, greve2024collaborative}. 3DSGs are particularly well-suited for capturing both geometric and semantic attributes, as nodes and edges can carry various properties, such as object affordances, spatial hierarchies, and semantic relationships~\citep{gu2024conceptgraphs, wu2021scenegraphfusion}. These representations have been shown to improve planning and reasoning capabilities, as demonstrated in systems like SayPlan~\citep{rana2023sayplan} and language-guided robots~\citep{honerkamp2024language}. While 3DSGs offer a rich framework for representing static environments, their ability to remain up-to-date in dynamic, real-world settings is limited. Most existing works focus on constructing and optimizing 3DSGs from static data, which can quickly become outdated in shared environments where objects and spatial layouts are constantly changing. This limitation is particularly problematic in non-stationary environments, as robots relying on outdated 3DSG representations are prone to errors in task execution. Our work addresses this gap by proposing MM-3DSGU, a framework that enables 3DSG updates using multimodal information sources, allowing robots to maintain a consistent and relevant understanding of their environment over time.

\textbf{Spatio-temporal Mapping}: To handle dynamic environments, spatio-temporal mapping methods have been developed to model object motion and environmental changes over time. Bore et al.\citep{bore2018detection} introduced a probabilistic approach to account for both short-term and long-term object motions within the environment. Building on this, Soares et al.\citep{virgolino2023visual} enhanced the robustness of ORB-SLAM3~\citep{ORBSLAM3_TRO} by filtering out dynamic points and applying a more reliable data association scheme to handle long-term dynamics. Similarly, Qian et al.\citep{qian2022pocd, qian2023pov} proposed an object-aware factor-graph formulation for SLAM that explicitly incorporates long-term dynamics of objects, allowing the map to evolve with changes in object locations. Khronos\citep{Schmid-RSS24-Khronos} further improves the handling of short-term dynamics by embedding objects' histories, ensuring that maps capture the temporal evolution of the environment. While these approaches provide valuable insights into dynamic mapping, they primarily focus on updating lower-level geometric maps. Our work extends beyond this by incorporating higher-level, semantic updates into 3DSGs, maintaining the consistency of the entire scene graph—both at the object and semantic relationship level—over time. This enables robots to not only account for spatial and geometric changes but also for semantic shifts, such as changes in object functions or relationships, which are critical for high-level reasoning.

\textbf{Change Detection}: Change detection is a key problem in dynamic environments, involving the identification of differences between two observations of the same scene~\citep{langer2020robust, park2021changesim, zhu2024living, sun2023nothing}. This problem is often addressed by comparing sensor data before and after changes occur to detect differences that warrant updates to the robot's internal map. Schmid et al.\citep{schmid2022panoptic} proposed using semantic consistency to detect changes and update the map accordingly, while Fu et al.\citep{fu2022planesdf, fu2022robust} leveraged the prior that objects tend to exist on planes, reducing the complexity of change detection in indoor environments. However, these approaches often focus on detecting changes in the geometric structure of the environment or object-level properties. Our framework, MM-3DSGU, expands upon these methods by incorporating multimodal inputs (such as temporal data, human interactions, and the robot's actions) to identify both geometric and semantic changes. This multimodal approach allows our system to not only detect when changes have occurred but also understand their context and relevance to the robot's task, ensuring that the updated scene graph is more aligned with the robot's current operational needs.

\subsection{Robot Learning in Non-Stationary Environments}

Robots operating in non-stationary environments must continually learn and adapt to changes in their surroundings. This has been an ongoing area of research in robotics, dating back to early work in Reinforcement Learning (RL) and Bayesian belief modeling, where robots attempt to learn policies that can generalize across varying environmental conditions~\citep{cox1994modeling, thrun1998bayesian, padakandla2020reinforcement}. However, there is no standardized approach for continual learning in non-stationary environments~\citep{graffieti2022continual, padakandla2021survey}, with existing methods often being task-specific or reliant on simplistic assumptions about the dynamics of the environment. Recent advances in lifelong learning for robots have focused on enabling adaptation in changing environments through techniques such as transfer learning, online learning, and meta-learning ~\citep{tan2018survey, hoi2021online, hospedales2021meta}. Yet, these approaches rarely address how scene representations themselves should be updated in real-time as the environment evolves. Our work fills this gap by proposing a structured update mechanism for 3D Scene Graphs, where the robot can dynamically modify its scene representation to reflect changes in the environment, ensuring continual alignment with real-world conditions. Moreover, unlike previous work that often assumes an explicit model of the environment’s dynamics, our framework leverages a flexible, multimodal approach that can generalize to a wide range of tasks and environments. By integrating diverse streams of information (e.g., visual perception, human-robot interactions, temporal context), we move toward a more generalisable solution for continual learning in dynamic settings, addressing a key limitation in existing literature.

\section{Problem Definition}

The problem we address can be framed as a \textbf{Partially Observable Markov Decision Process (POMDP)}~\citep{lauri2022partially, spaan2012partially}, where a robot operates in a dynamic environment represented by a \textbf{3D Scene Graph (3DSG)}. The 3DSG encodes both geometric and semantic information about objects and spaces, which the robot uses for reasoning and task execution. The environment is dynamic, with changes introduced by human activities and the robot’s own actions. At each time step, the robot receives partial observations from multiple modalities—such as human input, robot perception, and time—while it performs actions like exploring the environment or executing tasks. The robot’s challenge is to maintain an accurate scene graph by updating it as changes occur, ensuring its internal representation aligns with the real-world environment.

The core of the problem lies in defining an efficient \textbf{update function} $\mathcal{U}: O \times G_t \rightarrow G_{t+1}$, which takes the robot’s observations and the current scene graph and outputs an updated graph that reflects the latest changes. This multimodal approach integrates data from robot actions, human interactions, temporal semantics, and sensor inputs to detect and incorporate changes in both object locations and semantic relationships. The goal is to minimize discrepancies between the scene graph and the actual environment, ensuring that the 3DSG remains consistent and up-to-date, which in turn enables more effective task performance and decision-making in dynamic, non-stationary settings.

\section{Methodology}
\label{sec:methodology}
The proposed framework is agnostic to the change detection algorithms employed, provided that the algorithm adheres to the language and syntax needed to operate changes on the 3DSG through the usage of general primitives. In this work, we assume that the robot's pose is known, as in \citep{werby2024hierarchical}\citep{gu2024conceptgraphs}, and that an initial 3DSG has been built using any preferred state-of-the-art algorithm~\citep{hughes2022hydra}\citep{Maggio2024Clio}.

\subsection{Minimal Required 3D Scene Graph}
There have been numerous variations of 3DSGs~\citep{werby2024hierarchical}\citep{gu2024conceptgraphs}\citep{hughes2022hydra}, with differing hierarchical levels used to organize information and types of data embedded within the nodes.
In this work, the minimal granularity required consists of two hierarchical levels: the \textit{Object} and \textit{Room} layers. Depending on the extension of the workspace, it can be easily extended to include additional layers, such as \textit{Floors} and \textit{Buildings}, by applying the same logic used for the \textit{Room} layer. Here we focus on discussing the necessary information for the first two layers. In the \textit{Object} layer, as the name suggests, only object nodes are present. Each node contains the following information:
\begin{itemize}
    \item \textbf{Pose:} The 6DoF pose of the object \textit{o} with respect to the world \textit{w}, represented using the matrix $^{w}\mathbf{X}_{o} \in SE(3)$.
    \item \textbf{3D Bounding Box:} Minimal bounding box required to encapsulate the object \textit{o}, represented using the vector $\mathbf{b}_{o} \in R^{3}_{+}$, where $\mathbf{b}_{o} = (w, h, d)$, that are respectively the width, height and depth of the bounding box.
    \item \textbf{Label:} String that represents the semantic class of the object, represented with the vector $\textbf{l}_{o} \in L$, where $L$ is the set representing human language.
    \item \textbf{Decay Rate:} Parameter $\lambda_{o} \in R_+$ that controls the evolution of the likelihood that the object \textit{o} remains stationary after a time interval $\Delta t$. If the object is immovable, then $\lambda_\textit{o} = 0$.
    \item \textbf{Timestamp:} Value $t^o \in R_+$ representing when the object \textit{o} was last observed.
\end{itemize}
The \textit{Object} layer doesn't require edges between objects as opposed to \citep{gu2024conceptgraphs}. Only inter-layer edges, expressing the \textit{belonging} relation (i.e., an object \textit{o} belonging to a room \textit{r}), are used. In the \textit{Room} layer, as the name suggests, only room nodes are present. Each node contains a subset of the object node's elements: the pose $^{w}\mathbf{X}_{r} \in SE(3)$, 3D Bounding Box $\mathbf{r}_{o}$, and its label $\textbf{l}_{r}$. Additionally, room nodes are connected using intra-layer edges that express \textit{access to} relation(from one room to another).
 
\subsection{Muti-Modal Change Detection}
During the execution of a mission, the robot perceives, interacts with, and acts on the environment. While the research community has primarily focused on passive perception for map updates, we propose incorporating new potential sources of meaningful information, such as external agents and time. The Multi-Modal Change Detection module, one of our key contributions, mimics how a human would notice, be informed about, and predict changes in the environment in which they live or operate. Depending on the type of information received, the module will select the most appropriate change detection algorithm.

\textbf{\emph{Human Input}}: In a scenario where both robots and humans have access to the same spaces, the robot is no longer the only entity causing changes and moving objects. Humans are highly dynamic entities, and as such they will be responsible for most of the changes. Humans can acquire new objects, dispose of old ones, or move them from one location to another.
Thus, it would be extremely beneficial to leverage interactions with humans, for the robot to be informed of the changes that have occurred. With the advent of LLMs \citep{dubey2024llama}\citep{achiam2023gpt} and their high-level reasoning, processed text has become a reliable source of information.  We leverage the capabilities of LLMs to identify the type of operation, the object of interest \textit{o}, a possible supporting object \textit{so} (which could be used to disambiguate semantically similar objects), the source room \textit{sr} representing the location from which the object is moved or removed, and the target room \textit{tr} representing the location to which the object is moved or added. Humans communicate in a topological manner: ``I moved the cup from the kitchen to the table in the living room". In this example, which represents a common scenario, no accurate geometric indication object's placement is provided. Thus, when adding a new object or moving an existing one, the update operation will also be topological, involving a switch in the room to which the object belongs. The perception module will be then responsible for updating the node with accurate geometric information.

\textbf{\emph{Robot Action}}: If the robot lives in a shared environment with humans, we want it to assist us in our daily chores and tasks to make our lives easier. Although there is a wide variety of missions a human could assign to a robot, most share the same key components of interest: namely \emph{pick} and \emph{place}. These action primitives are the ones that cause changes in the environment. All the information needed to propagate these changes is found in the task description and in the task status. From the task description, we can infer the object \textit{o} of interest, along with the source room \textit{sr} and target room \textit{tr}. From the task status, we can determine when the object \textit{o} is being \emph{picked} (temporarily removed from the 3DSG) and \emph{placed} (re-attached to the 3DSG).

\textbf{\emph{Robot Perception}}: Most robots are equipped with cameras to perceive the external world similarly to humans. In this work, we assume the use of an RGB-D camera, allowing the robot to perceive simultaneously both visual and geometric features. To detect changes, the robot needs to compare the objects that it is expected to see with those currently perceived. Thus, the first step consists of querying the latest available version of the 3DSG for the list of objects that should be detected from its current pose $^{w}\mathbf{X}_{rb}^{t}$. 
We define the set of expected visible objects as $VO = \{ \textit{o} \in 3DSG \: \mid \: \textit{o} \in DO \wedge \textit{o visible from} \: ^{w}\mathbf{X}_{rb}^{t} \}$, where \textit{DO} is the set of all dynamic objects.
Next, we define the set of currently observed objects $BO$, where $\textit{bo} \in BO$ is represented as $\textit{bo} = (^{w}\mathbf{X}_{bo}, \mathbf{b}, \textbf{l})$. The semantic classes of the objects can be inferred using a closed-world object detector \citep{varghese2024yolov8}, while their poses and bounding boxes can be estimated using the point cloud obtained by projecting the depth values corresponding to the masks obtained from an object segmentation system \citep{kirillov2023segment}. The next step involves semantically and geometrically associating \textit{o} $\in VO$ and $\textit{bo} \in BO$. If both the semantic and geometric tests are passed, it signifies that \textit{o} has remained static. If only the semantic test is passed, then it is likely that \textit{o} has been moved to the location where \textit{bo} was observed. The non-associated objects in $VO$ are those we expected to observe but did not detect, making them candidates for removal from the 3DSG. Instead, the non-associated observed objects in $BO$  are those that were detected when they were not expected, making them candidates for being added to the 3DSG as new objects. To reduce data association complexity, we filter out the objects that are either immovable or unlikely to be moved, such as refrigerators and pantries.

\textbf{\emph{Time}}: Time is an additional source of information that has never been explicitly utilized to inform the robot of potential changes in the environment. As humans, we calibrate our expectations of how an environment evolves when we are not observing it for a period of time $\Delta t$, based on our prior semantic knowledge about persistency of objects in the environment and how others would interact with the various objects within that same environment. In this work, we simulate human's prior knowledge of how objects are interacted by utilizing the following function:
\begin{gather}
        p( d(o_{t}, o_{t_0}) < \epsilon \mid t, \textbf{l} ) = \frac{2}{1 + e^{\lambda_\textbf{l}(t - t_0)}}, \\
        d(o_{t}, o_{t_0}) = ||^{w}\mathbf{X}_{o}^{t} \boxminus ^{w}\mathbf{X}_{o}^{t_0} ||
\end{gather}
Where $d(o_{t}, o_{t_0}) = ||^{w}\mathbf{X}_{o}^{t} \boxminus ^{w}\mathbf{X}_{o}^{t_0} ||$ represents the distance between the poses of the same object when observed at times $t$ and $t_0$. $\epsilon$ represents the maximum displacement magnitude allowed for an object \textit{o} to still be considered static. The probability $p( d(o_{t}, o_{t_0}) < \epsilon \mid t, \textbf{l} )$ represents the likelihood that the object \textit{o} remains static after $t - t_0$, where $t_0$ is when it was last observed. The parameter $\lambda_\textbf{l}$ represents the decay rate of the function, controlling how rapidly the probability $p( d(o_{t}, o_{t_0}) < \epsilon \mid t, \textbf{l} )$ decreases. In other words, it indicates how quickly we expect an object to be interacted with by another human. This allows for a 'living' 3DSG which can evolve nodes on the graph as time passes.

Humans expect that the dynamics of different objects evolve differently based on their semantic class. To emulate this behaviour, we utilise LLMs to estimate the decay rate $\lambda_\textbf{l}$ of an object \textit{o} given the semantic class \textbf{l}. This choice stems from the general-purpose knowledge and semantic understanding of LLMs, which are derived from their training on internet-scale information.
$\lambda_\textbf{l} = 0.0$ indicates that the object is hardly moved or not moved at all. Each time an object \textit{o} is added, moved, or observed, the value $t_0$ is updated. Objects $\textit{o} \in 3DSG $ with $p( d(o_{t}, o_{t_0}) < \epsilon \mid t, \textbf{l} )$ below a certain threshold represent targets for the robot to actively perceive, as it becomes likely that these objects have been interacted with.


\subsection{Scene Graph Updater}
All these different modules provide a heterogeneous source of information. To establish a general framework that performs updates to the 3DSG, which is agnostic to the change detection algorithm used, it is necessary to define a \textit{Unified Language} to be used as the output of the change detection algorithm. It can be thought of as a dictionary with the following fields:
\begin{itemize}
    \item Source Room: Room's label $\textbf{l}_{r}$ in which the object \textit{o} belonged initially.
    \item Target Room: New room's label $\textbf{l}_{r}$ in which the object belongs to now.
    \item Target Object: Object's label $\textbf{l}_o$.
    \item Pose: Object's estimated pose $^{w}\mathbf{X}_{bo}$.  
    \item 3D Bounding Box: Object's estimated bounding box $\mathbf{b}_{o}$.
    \item Action: Three possible values that represent the outcome of the change detection algorithm: added/moved/removed. 
    \item Support Object: Semantic class of an object \textit{so} that has some topological relation with the object \textit{o} .
\end{itemize}
Using this information, the primitives for updating the 3DSG can be simplified to basic primitives that operate on any generic graph. The specific primitives and their sequence of use depend on the \textit{Action} field. The list of implemented primitives is the following:
\begin{itemize}
    \item Find($\textbf{l}$): Primitive used to find the node in the graph that corresponds to the input's label.
    \item Add(\textit{tr}, $\textbf{l}_o$, $^{w}\mathbf{X}_{o}$, $\mathbf{b}_{o}$): Primitive used to create a new object node and assign it to the target room \textit{tr}.
    \item Remove(\textit{sr}, $\textbf{l}_o$): Primitive used to remove an existing object node from its source room \textit{sr}.
    \item Move(\textit{sr}, \textit{tr}, $\textbf{l}_o$, $^{w}\mathbf{X}_{o}$): Primitive used to update the object's node information, which can be described as sequentially executing Remove and Add operations.  
\end{itemize}

\section{Experimental Setup}
\label{sec:experiments}

\textbf{Simulated Environment}:
To test the pipeline, a Unity environment has been created to enable control over all possible changes while also providing ground truth information about the objects within the environment. This environment consists of a single-story house with four rooms: a Bathroom, a Bedroom, a Living Room, and a Kitchen. Each room is populated with various sets of objects typically found in those spaces, for which we possess both CAD models and ground truth poses.

\textbf{Mission Specifications}: The robot, equipped with the MM-3DSGU module, is initially provided with a 3DSG built using the ground truth data available from the simulator before any changes are applied to the environment. A set of \textit{virtual actions}, simulating interactions a real human might have had with the environment, is then introduced. The goal, following the robot’s mission, is for it to detect all the changes made and accordingly update the 3DSG it was initially provided with.
The simulated mission is conducted as follows: the robot is informed by a human about a change they caused, such as, ``I ate the banana that was in the kitchen". It is then assigned a mission that begins with a \textit{pick} action and end with a \textit{place} action, for example, ``Pick the mug that is in the kitchen and take it to the bedroom." Throughout the mission, the robot passively monitors the environment for any changes that may have occurred.

The specifications of the simulated mission are as follows:
\begin{itemize}
    \item Human Action/Query: ``I removed the towel from the bathroom because it was too old."
    \item Mission: ``Pick the mug in the kitchen and take it to the bedroom."
    \item List of virtual actions: Removed ``banana", moved ``tv remote" from ``table" to ``sofa's arm", added ``book" on top of ``bed".
\end{itemize}

\textbf{Evaluation}: During the experimentation, the update operations performed on the 3DSG using MM-3DSGU are logged and annotated. For a preliminary and exploratory evaluation, these logged operations are compared with the real modifications, which were manually annotated. To ensure more fair and robust testing, the mission described was executed 10 times, and the reported results are averaged across those runs.

\section{Results}
\label{sec:result}

\tabref{table:prelim_res} summarizes the results obtained from the preliminary experiments. For each type of update operation, we measure the success rate and highlight the modules responsible for any occurred failures.
The \textit{Time} module was not evaluated due to its passive nature and the need for more rigorous formalization and metrics. The \textit{Action} module could never fail because grasping failures from the robot were not simulated. The \textit{RGB-D} module successfully recognized all the new objects introduced in the environment, but it incorrectly removed the ``tv remote" from the 3DSG, failing to recognize that it just had been moved to another location. YOLOv8 \citep{varghese2024yolov8} struggles to detect smaller objects, such as the ``tv remote", resulting in the incorrect removal of these items. GPT-4~\citep{achiam2023gpt} was employed for semantic association. The errors in failing to associate objects of the same semantic class and incorrectly associating objects of different semantic classes were mitigated by verifying that the associations were consistent during sequential observations. 
The \textit{Text} module success rate can be attributed to the topological nature of the updates, and the continuous improvement of GPT-4~\citep{achiam2023gpt}. The geometric attributes of added or moved objects will be updated by the \textit{RGB-D} module when perceived. The removal operation doesn't require any refinement.

\begin{minipage}[h]{\textwidth}
\vspace{5mm}
\centering
\parbox{0.96\textwidth}{\captionof{table}{Results of the framework obtained on the simulated scenario described above. The table showcases the different types of updates and the modules that were the cause of failure.}}
\vspace{-5mm}
\begin{longtable}{@{\extracolsep{\fill}}l rr rr rr rrrr @{}}

\toprule%
 & \multicolumn{2}{c}{{{\bfseries Success Rate}}}
 & \multicolumn{8}{c}{{{\bfseries Failure Rate}}}\\
 \cmidrule[0.4pt](r{0.125em}){1-11}

   \multicolumn{1}{c}{{{\bfseries Update Type}}}
 & \multicolumn{2}{c}{{{\bfseries }}}
 & \multicolumn{2}{c}{{{\bfseries Text}}}
 & \multicolumn{2}{c}{{{\bfseries RGB-D}}}
 & \multicolumn{2}{c}{{{\bfseries Action}}}
 & \multicolumn{2}{c}{{{\bfseries Time}}}\\

\cmidrule[0.4pt](r{0.125em}){1-1}
\cmidrule[0.4pt](lr{0.125em}){2-3}
\cmidrule[0.4pt](lr{0.125em}){4-5}
\cmidrule[0.4pt](lr{0.125em}){6-7}
\cmidrule[0.4pt](lr{0.125em}){8-9}
\cmidrule[0.4pt](lr{0.125em}){10-11}
\endhead

Add & 
\multicolumn{2}{c}{{{100.0\%}}} &
\multicolumn{2}{c}{{{0.0\%}}} &
\multicolumn{2}{c}{{{0.0\%}}} &
\multicolumn{2}{c}{{{0.0\%}}} &
\multicolumn{2}{c}{{{-}}} \\

\cmidrule[0.4pt](r{0.125em}){1-11}

Remove & 
\multicolumn{2}{c}{{{66.67\%}}} & 
\multicolumn{2}{c}{{{0.0\%}}} &
\multicolumn{2}{c}{{{33.33\%}}} &
\multicolumn{2}{c}{{{0.0\%}}} &
\multicolumn{2}{c}{{{-}}} \\
 
\cmidrule[0.4pt](r{0.125em}){1-11}

Move & 
\multicolumn{2}{c}{{{66.67\%}}} & 
\multicolumn{2}{c}{{{0.0\%}}} &
\multicolumn{2}{c}{{{33.33\%}}} &
\multicolumn{2}{c}{{{0.0\%}}} &
\multicolumn{2}{c}{{{-}}} \\
\bottomrule
\end{longtable}
\vspace{1mm}
\label{table:prelim_res}
\end{minipage}


\textbf{Future Work}: MM-3DSGU incorporates the concept of time into our object-centric representation, although this has yet to be explicitly exploited or evaluated. This research opens up the possibility of incorporating time into the robot's path planning to periodically check whether objects that are likely to have been moved remain static during the execution of other tasks. New metrics to assess the impact of this integration on overall mission performance still need to be formalized. Currently, the decay rate $\lambda$ is inferred using GPT-4~\citep{achiam2023gpt} from a set of discretized values, with each value associated with a \textit{representative} object serving as a basis for inferring  $\lambda$. These values were fine-tuned based on the authors' expectations regarding the dynamics of the selected elements. While this decay-tuning serves as a proof-of-concept, it would be beneficial to develop a more general and informed method for setting the parameter $\lambda$ or to incorporate learning for adapting these parameters to the specific environments where the robot operates. Further experimentation in both simulation and real-world settings is needed to effectively validate the system performance. This work highlights the necessity for new types of annotations in datasets, such as temporal data, especially if the robot is expected to share workspaces with humans, along with the need for new metrics to evaluate these performances.

\section{Conclusion}
\label{sec:conclusion}

In this work, we presented preliminary insights about MM-3DSG, the first general framework for updating a 3DSG that combines human interactions, robot actions, and perception while accounting for the evolution of the environment over time. To address heterogeneous sources of information, we introduced a common language and primitives for unifying and generalizing update operations on a 3DSG. The initial simulated experimentation indicates that this approach shows promising results, opening up various new research directions.

\clearpage
\acknowledgments{I would like to thank my friend and colleague, Nicolas Chapman, for always being open to discussion and for helping me navigate various challenges.}


\bibliography{references}  

\end{document}